\documentclass[final]{cvpr}
\usepackage{cuted}
\usepackage{capt-of}
\usepackage{times}
\usepackage{epsfig}
\usepackage{graphicx}
\usepackage{enumitem}
\usepackage{amsmath}
\usepackage{amssymb}

\usepackage[pagebackref=false,breaklinks=true,letterpaper=true,colorlinks,urlcolor=blue,citecolor=blue,linkcolor=blue,bookmarks=false]{hyperref}
\def\swsix{0.14\linewidth}
\def\swseven{0.14\linewidth}
\def\cvprPaperID{6187} % *** Enter the CVPR Paper ID here
\def\confYear{CVPR 2021}
\def\sweight{0.124\linewidth}

\begin{document}
\newcommand{\figdir}{figures}
%%%%%%%%% TITLE
% \title{PD-GAN: A Probablistic View of Diverse Image Inpainting}
\title{PD-GAN: Probabilistic Diverse GAN for Image Inpainting}

\author{Hongyu Liu$^1$ \quad
Ziyu Wan$^{2}$ \quad
Wei Huang$^{3}$ \quad
Yibing Song$^4$ \quad
Xintong Han$^1$\thanks{X. Han is the corresponding author. The results and code are available at \url{https://github.com/KumapowerLIU/PD-GAN}.}  \quad
Jing Liao$^2$ \\
$^{1}$Huya Inc \quad
$^{2}$City University of Hong Kong \quad
$^{3}$Hunan University \quad $^{4}$Tencent AI Lab\\
{\tt\small \{liuhongyu1,hanxintong\}@huya.com \quad ziyuwan2-c@my.cityu.edu.hk}\\
{\tt\small  yibingsong.cv@gmail.com \quad jingliao@cityu.edu.hk}

}
% %

% \renewcommand*{\Affilfont}{\small\it} %
% \renewcommand\Authands{ and }
% \date{}

\def\swsix{0.132\linewidth}
\def\swseven{0.14\linewidth}
\def\sweight{0.124\linewidth}

\maketitle
\pagestyle{empty}  % no page number for the second and the later pages
\thispagestyle{empty} % no page number for the first page
% before \begin{abstract}
\begin{strip}
\vspace{-0.5in}
\setlength\tabcolsep{0.5pt}
\centering
\begin{tabular}{cccccccc}
    \vspace{-0.5mm}
    \includegraphics[width=\swseven]{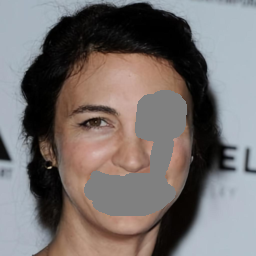}&
    \includegraphics[width=\swseven]{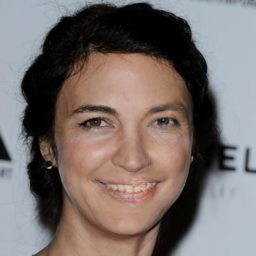}&
    \includegraphics[width=\swseven]{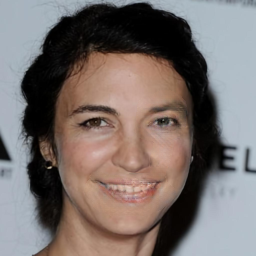}&
    \includegraphics[width=\swseven]{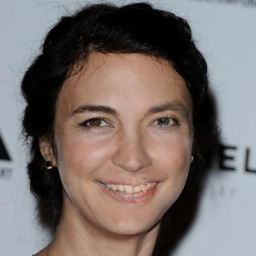}&
    \includegraphics[width=\swseven]{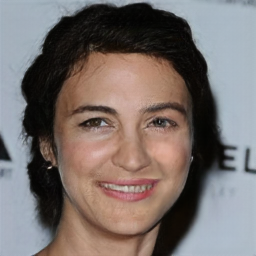}&
    \includegraphics[width=\swseven]{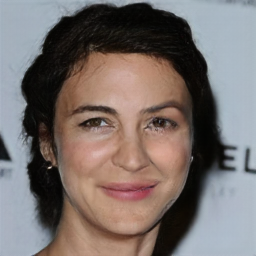}&
     \includegraphics[width=\swseven]{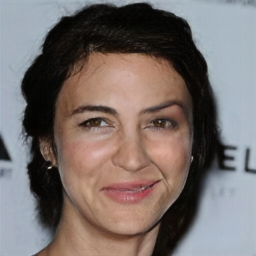}\\
    \vspace{-0.5mm}
    \includegraphics[width=\swseven]{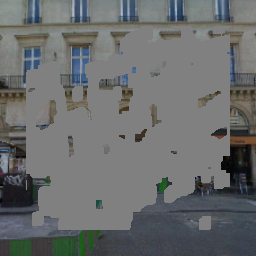}&
    \includegraphics[width=\swseven]{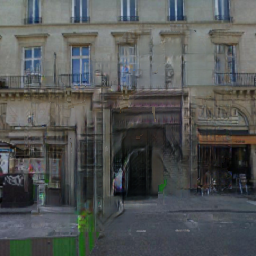}&
    \includegraphics[width=\swseven]{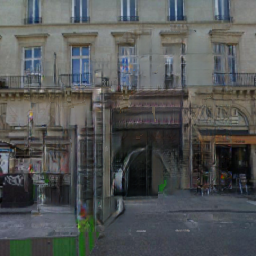}&
    \includegraphics[width=\swseven]{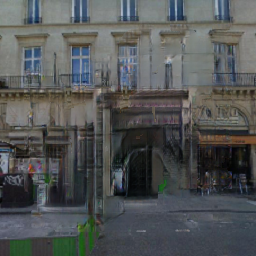}&
    \includegraphics[width=\swseven]{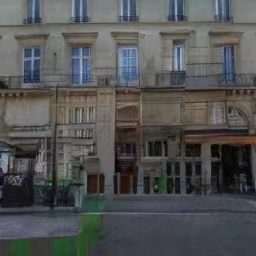}&
    \includegraphics[width=\swseven]{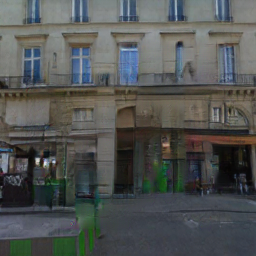}&
    \includegraphics[width=\swseven]{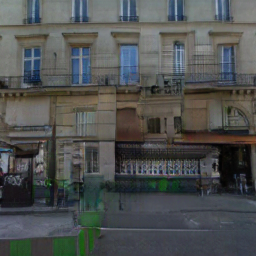}\\
    \vspace{-0.5mm}
    \includegraphics[width=\swseven]{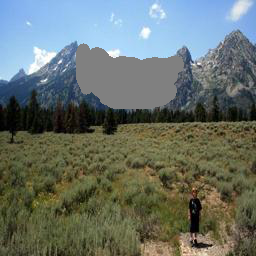}&
    \includegraphics[width=\swseven]{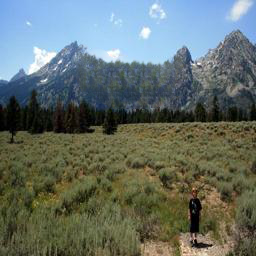}&
     \includegraphics[width=\swseven]{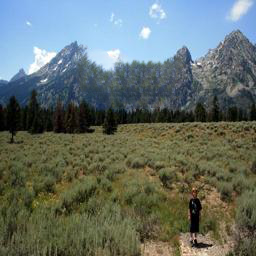}&
     \includegraphics[width=\swseven]{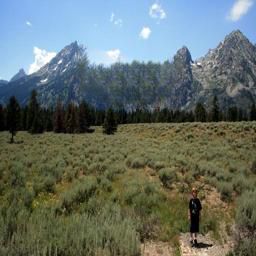}&
     \includegraphics[width=\swseven]{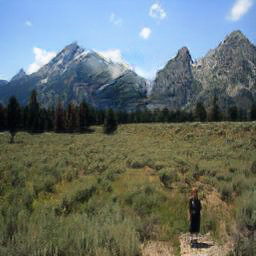}&
      \includegraphics[width=\swseven]{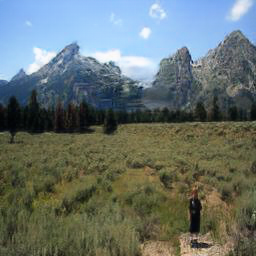}&
      \includegraphics[width=\swseven]{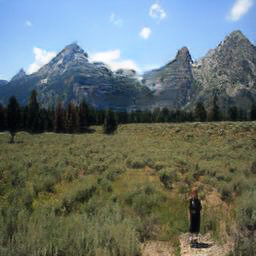}\\

   (a) Input  &  (b) PIC$_1$~\cite{zheng2019pluralistic}  & (c) PIC$_2$  &(d) PIC$_3$  &(e) Ours$_1$   & (f) Ours$_2$  & (g) Ours$_3$
     \end{tabular}
      \vspace{1mm}
\captionof{figure}{Visual comparison of diverse inpainting results. Our method generates more diverse and visually realistic results than PIC~\cite{zheng2019pluralistic}.
%The input images are shown in (a), the generated results of PIC~\cite{zheng2019pluralistic} are shown in (b)-(d), the results of our method are shown in (e)-(g). In contrast to PIC~\cite{zheng2019pluralistic}, our method generates more diverse and higher quality results.
}
\label{fig:teaser}
\end{strip}

%%%%%%%%% ABSTRACT
\begin{abstract}

We propose PD-GAN, a probabilistic diverse GAN for image inpainting. Given an input image with arbitrary hole regions, PD-GAN produces multiple inpainting results with diverse and visually realistic content. Our PD-GAN is built upon a vanilla GAN which generates images based on random noise. During image generation, we modulate deep features of input random noise from coarse-to-fine by injecting an initially restored image and the hole regions in multiple scales. We argue that during hole filling, the pixels near the hole boundary should be more deterministic (\ie, with higher probability trusting the context and initially restored image to create natural inpainting boundary), while those pixels lie in the center of the hole should enjoy more degrees of freedom (\ie, more likely to depend on the random noise for enhancing diversity). To this end, we propose spatially probabilistic diversity normalization (SPDNorm) inside the modulation to model the probability of generating a pixel conditioned on the context information. SPDNorm dynamically balances the realism and diversity inside the hole region, making the generated content more diverse towards the hole center and resemble neighboring image content more towards the hole boundary. Meanwhile, we propose a perceptual diversity loss to further empower PD-GAN for diverse content generation. Experiments on benchmark datasets including CelebA-HQ, Places2 and Paris Street View indicate that PD-GAN is effective for diverse and visually realistic image restoration.

\end{abstract}

% inpainting task
\section{Introduction}
There is a growing attention on developing advanced image inpainting methods for content removal~\cite{shetty2018adversarial,Barnes-sig09-patchmatch} and image restoration~\cite{song2019joint,wan2020bringing,wang2020rethinking,wan2020old}. Based on deep CNNs, image inpainting methods~\cite{Pathak-cvpr16-featinpaint,liu2020rethinking,Yu-cvpr18-attentioninpainting,liu-2019iccv-coherent} typically utilize an encoder-decoder network to generate  meaningful image content for hole filling. Meanwhile, the content across the hole boundary is enforced consistent during visually realistic generation. By taking the input image with hole regions, the encoder captures deep image representations hierarchically, which are decoded to produce the output result.

%Image inpainting refers to utilize the context information to fill the corrupted regions for visual aesthetics improvement. This task is a fundamental problem in the field of image manipulation and can be used in various applications such as unwanted object removal \cite{shetty2018adversarial,Barnes-sig09-patchmatch3} and photo restoration \cite{wan2020bringing}. The key challenge of inpainting is not only to generate natural and reasonable content for missing regions, but also to ensure the coherence between filled regions and known visible contextual.
% traditional transition to learning
%Recently, deep learning with generative adversarial networks (GANs) have been used to solve the inpainting task. Unlike the traditional methods~\cite{Efros-sig01-quilting1,Barnes-sig09-patchmatch3,Efros-ICECCS01-synthesis2,bertalmio-SIG2000-imageinpainting18} that propagate background image content to the hole regions via patch-based image matching, deep inpainting methods~\cite{Pathak-cvpr16-featinpaint7,Iizuka-sig17-lgmatch4, Iizuka-sig17-lgmatch4,Li-cvpr17-facecomplete5,Yeh-arxiv1607-spinpaint6,Pathak-cvpr16-featinpaint7,liu2020rethinking,Yu-cvpr18-attentioninpainting8,liu-2019iccv-coherent44} model the inpainting task as encoder-decoder framework and produce more meaningful and globally consistent results.
%However this framework suppresses the creativity of GANs and can only generate only one optimal result since the input of decoder is unique which is obtained by image encoding.

While deep encoder-decoders improve the image inpainting performance, they target for single image generation that each input image corresponds to one restored result. In practice, image inpainting may produce multiple results because of the uncertain content generation within the hole region. This diverse image inpainting is less touched by existing inpainting methods. Recently, investigations~\cite{zheng2019pluralistic, zhao2020uctgan, han2019finet} on diverse image inpainting follow the encoder-decoder structure. They are not effective to generate both diverse and realistic contents. One reason is that these methods still utilize the encoder to model the current masked image to Gaussian distribution and decode to a completed image, the variation of distribution is greatly limited by the masked image itself which leads to a decline in diversity, especially when the hole regions are free-form. On the other hand, these encoder-decoder networks utilize image reconstruction loss \cite{Justin-eccv16-perceptual} during training. The generated content is thus enforced to be similar to the ground truth across both low-level and semantic representations. Heavily relying on such reconstruction loss limits the diverse content generation.

In this work, we propose PD-GAN, a diverse image inpainting network built upon a vanilla GAN. We notice that GAN is powerful to generate diverse image content based on different random noise inputs. Thus, instead of sending input images to the CNN, our PD-GAN starts from a random noise vector and then decodes this noise vector for content generation. In all the decoder layers, we inject prior information (coarse reconstruction result from a pre-trained partial convolution model \cite{liu-2018eccv-particalcov}) and the region mask. The injection is fulfilled by the proposed SPDNorm (spatially probabilistic diversity normalization) module. SPDNorm gradually modulates deep features of the noise vector with input image representations. Specifically, the SPDNorm module learns a spatial transformation containing both hard and soft probabilistic diversity maps for feature fusion. The diversity is enhanced towards the hole center while is reduced towards the hole boundary.

Moreover, we propose a perceptual diversity loss to empower the diverse generation ability of PD-GAN. For two output images generated by the same prior information but input noise vectors, the perceptual diversity loss forces these two images to be farther in feature space. By training PD-GAN with the perceptual diversity loss, we can effectively generate both diverse and visually realistic contents for image inpainting. Some results can be found in Fig~\ref{fig:teaser}.

Our contributions are summarized as follows:
\begin{itemize}[noitemsep,nolistsep]
\item Based on a vanilla GAN, the proposed PD-GAN modulates deep features of random noise vector via the proposed SPDNorm to incorporate context constraint.
\item We propose a perceptual diversity loss to empower the network diversity.
\item Experiments on the benchmark datasets indicate that
our PD-GAN is effective to generate diverse and visually realistic contents for image inpainting.
\end{itemize}

% $\bullet$ We propose the spatially probabilistic diverse normalization to guarantee the diversity of predictions while ensuring appearance and structure consistency in the whole image.

%-------------------------------------------------------------------------
% pipeline
\begin{figure*}[t]
\begin{center}
\includegraphics[width=1.0\linewidth]{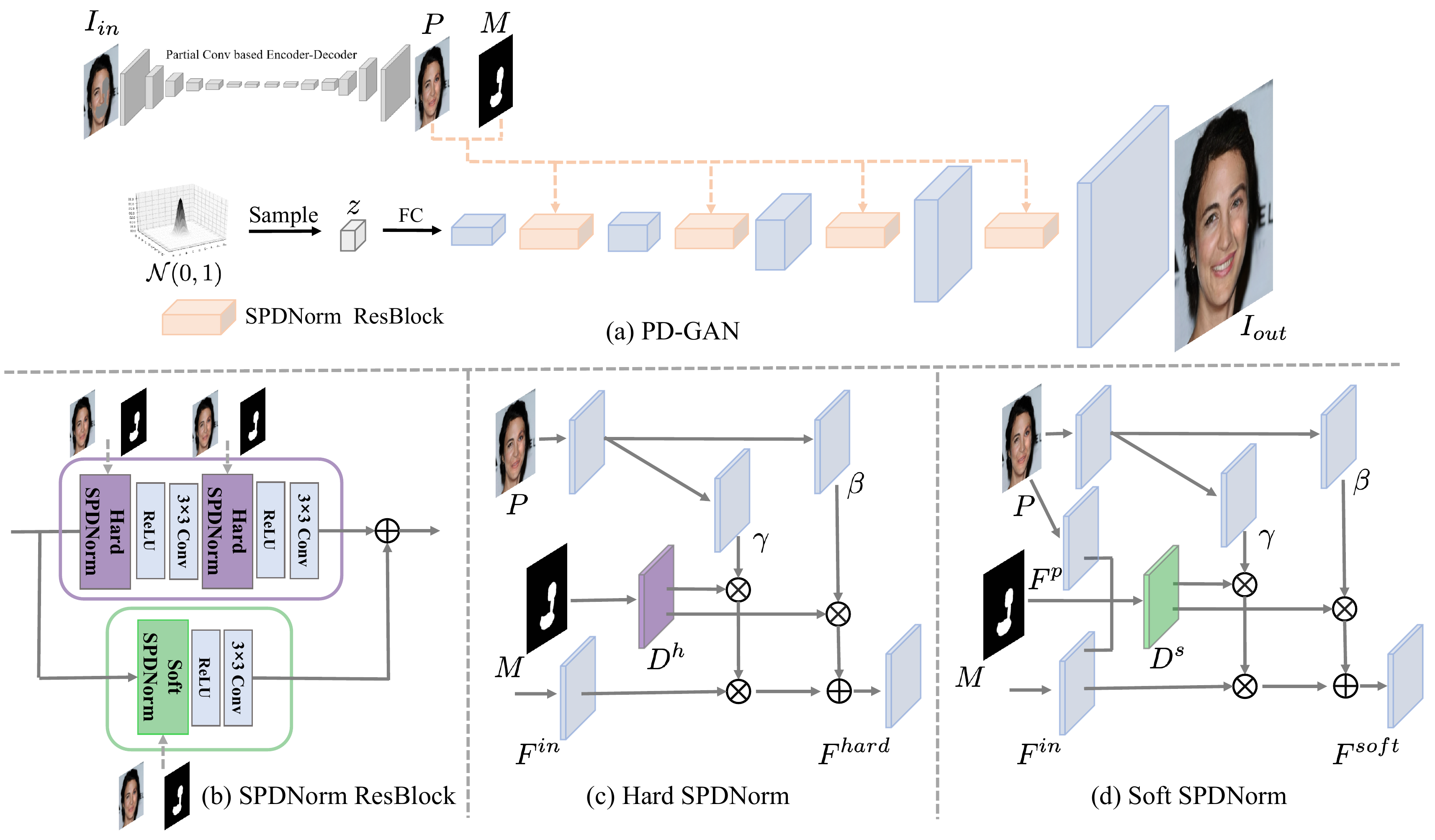}
\caption{An overview of PD-GAN with SPDNorm. (a) We first get a coarse prediction as the prior information from a pre-trained partial conv~\cite{liu-2018eccv-particalcov} based network. Then we sample the latent vector $z$ from a standard Gaussian distribution and PD-GAN modulates $z$ with the prior information based on the SPDNorm Residual block. (b) The SPDNorm Residual block consists of hard and soft SPDNorm. (c-d) The hard and soft SPDNorm control the confidence of prior information based on the mask itself and a learning process, respectively.}
\label{fig:pipeline}
\end{center}

\end{figure*}

%-------------------------------------------------------------------------
% Empirical, single solution, diverse, generation  related work

\section{Related Work}
{\flushleft \bf Image Inpainting.}
Existing inpainting methods can be divided into two categories:  single-solution inpainting methods and diverse inpainting methods. The single-solution inpainting methods produce a single result for each masked image, while the diverse inpainting methods generate multiple  results for each corrupted image.

\textit{Single-solution inpainting methods:}
Some traditional single-solution image inpainting methods~\cite{bertalmio-SIG2000-imageinpainting,levin-2003-learning,ballester-2001tio-filling} based on diffusion techniques propagate the contextual appearances to the missing regions. Other methods~\cite{criminisi-2004tip-region,darabi-2012tog-imagemelding,song2017stylizing,Barnes-sig09-patchmatch,xu-2010-image}  based on patch match fill missing regions by calculating the statistics of patch offsets and transferring similar patches from the undamaged region to the hole region. The deep learning based  single-solution inpainting methods~\cite{Pathak-cvpr16-featinpaint, Iizuka-sig17-lgmatch, yu-2015-multidilated, nazeri-2019ICCVW-edgeconnect,  xiong-2019CVPR-foreground, ren-2019iccv-structureflow,Yu-cvpr18-attentioninpainting, Song-eccv18-iftinpainting,liu-2018eccv-particalcov,Yan-eccv18-shiftnet,yu-2019iccv-freeform,liu-2019iccv-coherent,wang-2018NIPS-imageINPAINTING,liu2020rethinking,Yi_2020_CVPR,dong2020fashion,Dong_2019_ICCV,Han_2018_CVPR}  typically
involve the generative adversarial networks~\cite{goodfellow-2014nips-generative} to learn the semantic of image. These single-solution inpainting methods achieve high performance in predicting deterministic result for the hole regions, but they cannot generate a variety of semantically meaningful results.

\textit{Diverse inpainting methods:} To generate pluralistic results given a corrupted image, \cite{zhao2020uctgan,zheng2019pluralistic,han2019finet} %utilize the KL-Divergence loss term~\cite{kingma2013auto} to condition the image generation on a Gaussian distribution, then they sample the latent vector $z$ from the distribution and decode the $z$ to the image during training process.
train conditional VAE \cite{walker2016uncertain} type of encoder-decoder networks. They encode the masked image to condition a Gaussian distribution, from which stochastic sampling at the test time achieves diverse inpainting results. However, the degree of diversity is controlled by the dispersion of distribution, and the dispersion is limited by the masked image. In contrast, our method samples a latent vector from the standard Gaussian distribution and map the latent vector to image directly by a single decoder. Meanwhile, these methods mainly rely on the image reconstruction loss during training, but unconditionally forcing the results similar to the ground truth will decrease the diversity of outputs. We propose the perceptual diversity loss to handle this issue.

{\flushleft \bf Diverse image generation methods.}
In addition to diverse inpainting methods,  the diverse image generation methods, such as VAE~\cite{kingma2013auto},  GANs~\cite{goodfellow-2014nips-generative} and CVAE~\cite{walker2016uncertain}, can also generate the diverse results.
BicycleGAN~\cite{zhu2017toward} makes invertible connections between the latent code and the generated image,  which helps produce diverse results. MUNIT~\cite{huang2018multimodal} and EBIT~\cite{Zhang_2020_CVPR} combine the content and style from different images to achieve diverse image-to-image translation. However, these methods are not designed  for image inpainting task, so the inpainted results usually present artifacts.

{\flushleft \bf Normalization Layers.}
There are wide investigations of the normalization layers~\cite{huang2017arbitrary,salimans2016weight,ioffe2015batch,wu2018group,ba2016layer,ulyanov2016instance,song2017crest,li2019positional} in deep learning to improve the network prediction performance. Among them, Spatially-adaptive denormalization (SPADE)~\cite{park2019semantic} relates to our SPDNorm in that SPADE is a simple form of SPDNorm. By fully relying on the prior information, SPADE is not effective to empower network for diverse content output.

%The SPADE is a special variant of our proposed SPDNorm (SPDNorm abandon the probabilistic diversity maps).  The SPADE trust the inject information completely  result in the unstable training and model collapse. This is because too much guide information injection will go against diversity.

% the They apply spatially-varying affine transformations which make the generative result refer to the semantic mask. However, SPADE fails to handle diverse image inpainting task. This is because the styles and semantics cannot be disentangled in a single masked image, and the semantic mask is hard to get for a corrupted image.

 %-------------------------------------------------------------------------

%-------------------------------------------------------------------------
\section{PD-GAN}

\label{sec:PD-GAN}
Fig.~\ref{fig:pipeline} shows an overview of probabilistic diverse generative adversarial network (PD-GAN), which sets the coarse result as prior information and modulate the latent vector $z$ to image space by a single decoder similar to a vanilla GAN. We utilize the pre-trained Partial Convolutional encoder-decoder~\cite{liu-2018eccv-particalcov} to get the coarse prediction. The coarse prediction and mask image are sent to SPDNorm Residual Blocks to provide prior knowledge for the generation process. The SPDNorm Residual Block consists of SPDNorm with hard and soft probabilistic diversity maps (Hard SPDNorm and Soft SPDNorm). The pixels  close  to  the  hole  boundary  should  be  more  deterministic and the probability of generating diverse result is small, while those pixels at the center of the hole should enjoy more degrees of freedom and the probability of generating diverse result is large. The hard SPDNorm controls the probability  according to the distance between the pixel and hole boundary, while the soft  SPDNorm learns the probability in an adaptive process. In this section, we first describe SPDNorm in detail. % In this section,  we first introduce the SPDNorm, then we analyze the characteristics of hard and soft SPDNorm. Finally, we adopt a residual block combining the advantages of both forms to achieve better performance.

% {\flushleft \bf Motivation}
% The previous diverse image inpainting methods utilize a variant of image to image framework to generate multiple results. These methods typically  encode the masked image to a distribution which is close to the ground truth distribution, then randomly sample latent vector from the distribution and decode this vector to image. However, the variation in distribution is limited by the masked image, and the degree of numerical change in the latent vector is limited at the beginning. So the diversity of generated results is not good. To tackle this issue, we sample latent vector $Z$ from $\mathcal N (0,1)$ and decode the vector to image by single decoder and borrowing some information from prior prediction. Our method do not limit the variation of $Z$ at the beginning, but modulate the $Z$ progressively by the SPDNorm.
% SPDNORM
\subsection{SPDNorm}

\begin{figure}[t]
\begin{center}

\begin{tabular}{cc}
    \vspace{-0.5mm}
  \includegraphics[width=0.80\linewidth]{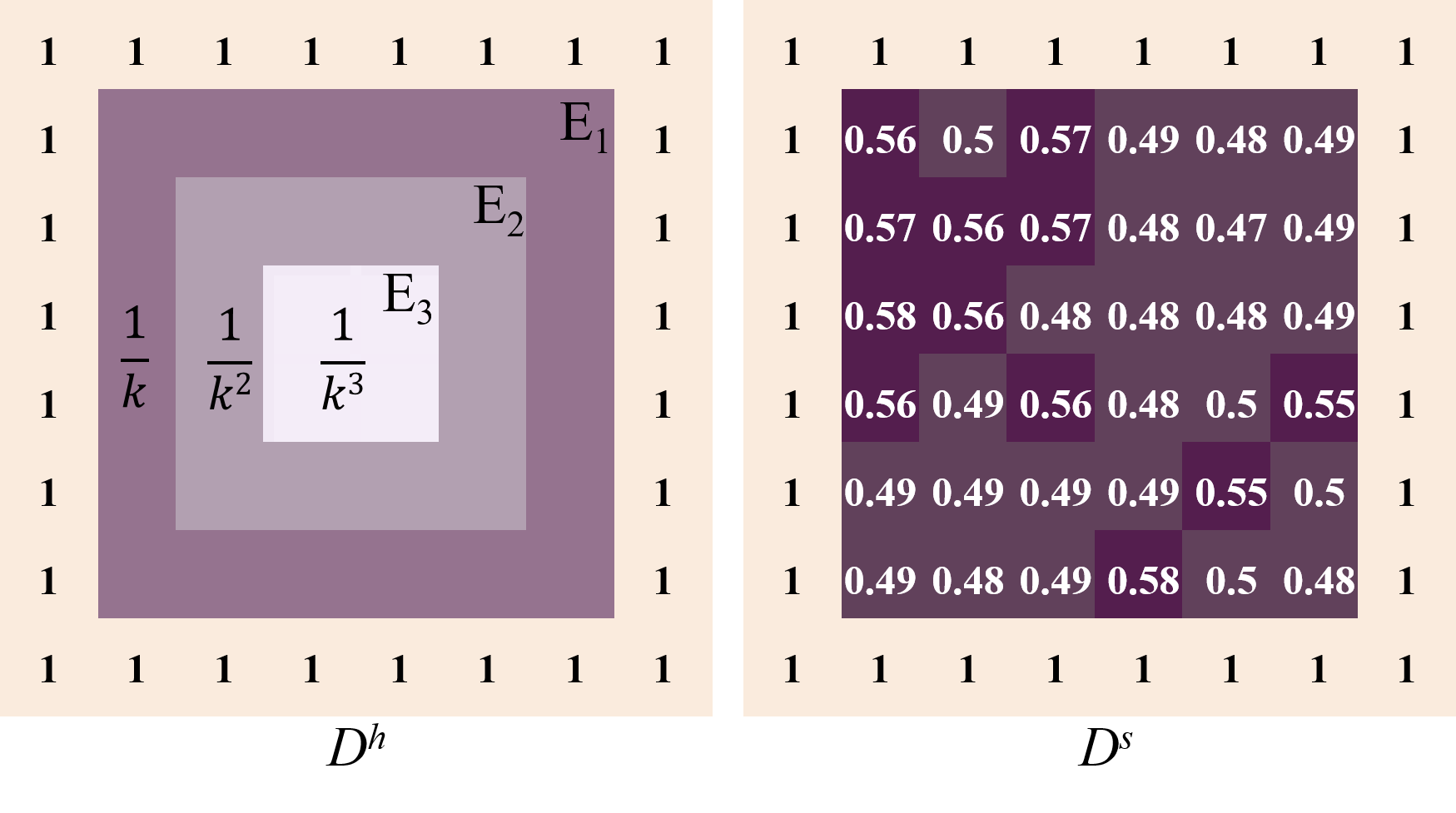}

\end{tabular}
\caption{An example of hard probabilistic diversity map $D^h$ and soft probabilistic diversity map $D^s$. The area with a value of 1 is the background. We show $D^h$ and $D^s$ at the stage with the feature map size of $8\times8$. The value of $D^h$ gradually decreases from boundary  to center. The value of $D^s$ changed adaptively with a learning process. }
\label{fig:mask}
\end{center}
\end{figure}

The SPDNorm learns the scale and bias to transform the feature map. It contains hard and soft SPDNorm layers that are controlled by hard and soft probabilistic diversity maps, respectively. The hard probabilistic diversity map $D^h$ is determined by the inpainting mask $M$ without learning process. The soft probabilistic diversity map $D^s$ is an adaptive map which is obtained by the input feature and coarse prediction with a learning process. As shown in Figure \ref{fig:pipeline}(c-d), We denote $F^{\rm in} \in \mathbb{R} ^ {H\times W\times C}$ as the input feature map and $P \in \mathbb{R} ^ {H\times W\times 3}$ as the prior information (coarse prediction). We denote the outputs of hard and soft SPDNorm as $F^{hard}$ and $F^{soft}$, respectively:
\begin{equation}\label{eq:hspd}
   F^{hard}_{x,y,c} =D^h_{x, y}(\gamma_{x, y, c}(P)\frac{F^{\rm in}_{x,y,c}-\mu_{c}}{\sqrt{\sigma_{c}^{2}+\epsilon}}+ \beta_{x, y, c}(P)),
\end{equation}
\begin{equation}\label{eq:sspd}
   F^{soft}_{x,y,c} =D^s_{ x, y}(\gamma_{x, y, c}(P)\frac{F^{\rm in}_{x,y,c}-\mu_{c}}{\sqrt{\sigma_{c}^{2}+\epsilon}}+ \beta_{x, y, c}(P)),
\end{equation}
where $\gamma_{x, y, c}(P)$ and $\beta_{x, y, c}(P)$ are two variables  output by two convolutional layers to control the element-wise influence from the coarse prior information $P$. $\mu_{c} = \frac{1}{H\times W} \sum_{x=1}^{H} \sum_{y=1}^{W} F^{\rm in}_{ x, y, c}$,  $\sigma_{c}=\sqrt{\frac{1}{H\times W} \sum_{x=1}^{H} \sum_{y=1}^{W}\left(F^{\rm in}_{x,y,c}-\mu_{c}\right)^{2}}$. % In practice, we use two convolutional layers to generate $\gamma_{x, y, c}(P)$ and $\beta_{x, y, c}(P)$ at each element location. The
  Fig.~\ref{fig:mask} shows an example of $D^h$ and $D^s$.

 \renewcommand{\tabcolsep}{0.01pt}

{\flushleft \bf Hard SPDNorm.}
Intuitively,  the closer the content is to the boundary, the stronger the constraints of the context. The content that is close the hole boundary needs to be consistent with the context, and thus we need more prior information to guide the  hole filling process. While for the regions close to the hole center, PD-GAN needs less prior information and  has a higher probability to generate diverse results. We control the probability by a hard probabilisitc diversity map $D^h$ as shown in Fig.~\ref{fig:mask}. For the mask $M$ (0 for the missing region, and 1 for background), we apply $n$ iterative dilation operations to it. The dilation is realized by a mask update process~\cite{liu-2018eccv-particalcov}.  Specifically, we denote the mask update process as $F_m$. The mask after the $i$-th dilation operation is $M_i$, which is obtained by applying the update process on the mask from previous step (\ie, $M_i=F_m(M_{i-1})$ with $M_0=M$). Mathematically, the mask update process $F_m$ can be expressed as:
\begin{equation}
\label{eq:maskprocess}
M_i(x, y)=
\begin{cases}
% 1&  \text{if sum(}m_{(a, b)} \text{)!=0} \\
1&  \text{if ~} \sum_{(a, b) \in \mathcal{N}(x,y)} M_{i-1}(a, b) > 0 \\
0& \text{otherwise}
\end{cases}    ,
\end{equation}
where $\mathcal{N}(x,y)$ denotes the $3\times3$ neighborhood region centered at location $(x, y)$. Finally, we fill in the $i$-th dilated region $E_i=M_i-M_{i-1}$ in $D^h$ with value $\frac {1}{k^{i}}$ as shown in Fig.~\ref{fig:mask}. We empirically set $k=4$ in our paper.

As a result, we can get $D^h$ whose value in the hole regions is exponentially decreases as the location is closer to the hole center, and is 1 outside hole regions (\ie, background). As the value decreases, the credibility of prior information becomes lower, and the probability of diversity increases. There is no learning mechanism in the generation of $D^h$ (\ie, $D^h$ is fixed). Our hard SPDNorm cannot only intuitively make the generated content and the background area coherent, but also ensure the diversity of the prediction.

Note that we set the number of mask updates to $n = 2, 2, 4, 4, 4$ for each stage from deep to shallow layers as the resolution of input mask increases.

{\flushleft \bf Soft SPDNorm.}
The Hard SPDNorm explicitly controls the probability of generating diverse results  according to the $D^h$. However, this probability should also be dynamic, which depends on the prior information and the inpainting mask. In other words, our network should be learnable and have the ability of paying attention to certain regions conditioned on the prior information and the inpainting mask. To this end, we propose soft SPDNorm to adaptively learn the probability of producing multiple content to achieve better inpainting results. The soft SPDNorm extracts the feature from both  $P$ and $F^{\rm in}$ to predict a soft probabilistic diversity map, guiding the diverse inpainting process. As Fig. \ref{fig:pipeline}(d) illustrates, we extract the feature map $F^p$ from prior information $P$ by convolutional layers, then $F^p$ and input feature map $F^{\rm in}$ are concatenated to get $D^s$:
\begin{equation}
   D^s = \sigma (\operatorname{Conv}([F^p, F^{\rm in}])\cdot(1-M)+M,
\label{eq:smask}
\end{equation}
where $\sigma$ is the sigmoid activation function and the elements corresponding to background in $D^s$ are set to $1$. In order to achieve stable training and generate plausible results, $D^s$ adaptively changes probability of borrowing information from $P$. We find that the value of $D^s$ in the foreground region changes smoothly and is close to $0.5$, so only relying on $D^s$ is unable to measure probability of predicting diverse results .

{\flushleft \bf SPDNorm ResBlock.}
As discussed above, the hard SPDNorm increases the probability of getting diversity results but reduces the quality of the results. In contrast, the soft SPDNorm can stabilize the training and dynamically learn the condition of the prior information but lack diversity. So we propose the SPDNorm ResBlock to let them complement each other as shown in Fig.~\ref{fig:pipeline}. Meanwhile, note that each residual block operates at a different scale, so we downsample the prior information and mask to match the corresponding spatial resolution.

\renewcommand{\tabcolsep}{0.5pt}
\begin{figure*}[t]
    \begin{center}
    \small
\begin{tabular}{cccccc}
\vspace{-0.5mm}
\includegraphics[width=\swsix]{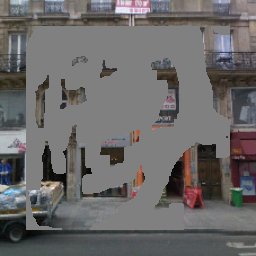}&
\includegraphics[width=\swsix]{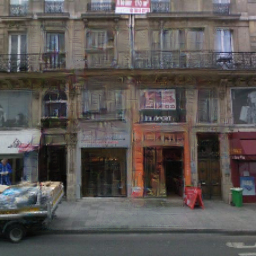}&
\includegraphics[width=\swsix]{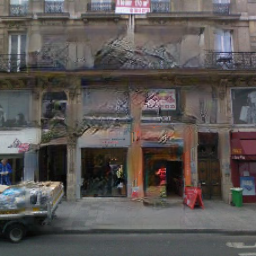}&
\includegraphics[width=\swsix]{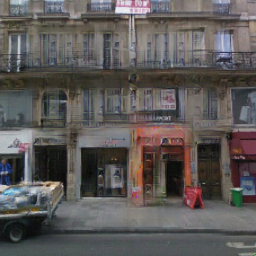}&
\includegraphics[width=\swsix]{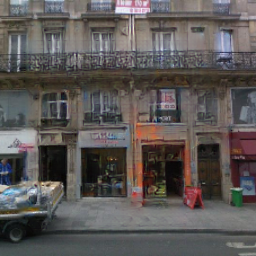}&
\includegraphics[width=\swsix]{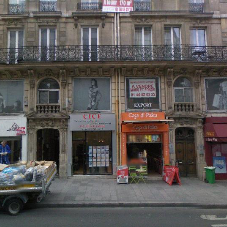}\\
(a) Input & (b) EC~\cite{nazeri-2019ICCVW-edgeconnect} &(c) GC~\cite{yu-2019iccv-freeform} &(d) PC~\cite{liu-2018eccv-particalcov}   & (e)  RFR~\cite{Li_2020_CVPR} & (f) GT \\
\includegraphics[width=\swsix]{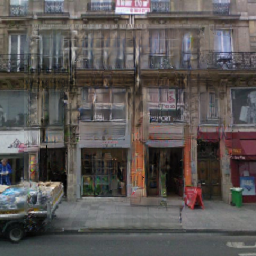}&
\includegraphics[width=\swsix]{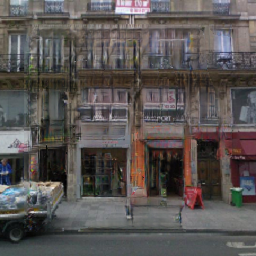}&
\includegraphics[width=\swsix]{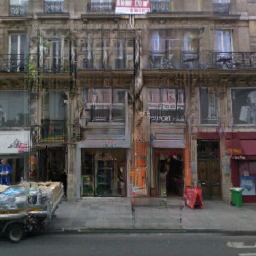}&
\includegraphics[width=\swsix]{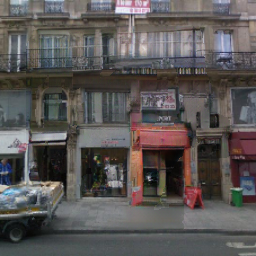}&
\includegraphics[width=\swsix]{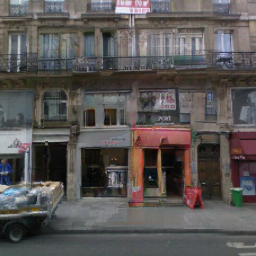}&
\includegraphics[width=\swsix]{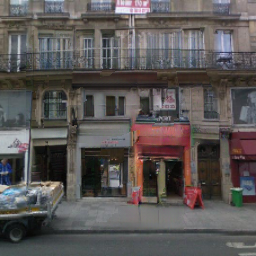}\\
(g) PIC$_1$~\cite{zheng2019pluralistic}  & (h) PIC$_2$  &(i) PIC$_3$  &(j) Ours$_1$   & (k) Ours$_2$  & (l) Ours$_3$  \\
\includegraphics[width=\swsix]{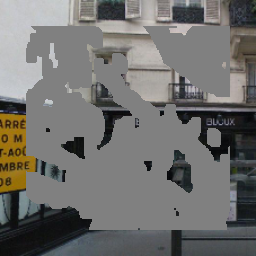}&
\includegraphics[width=\swsix]{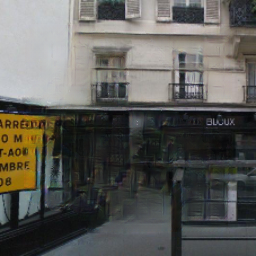}&
\includegraphics[width=\swsix]{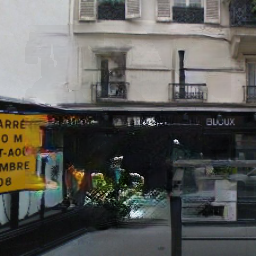}&
\includegraphics[width=\swsix]{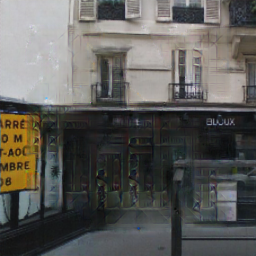}&
\includegraphics[width=\swsix]{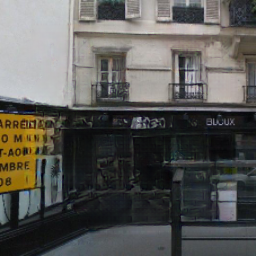}&
\includegraphics[width=\swsix]{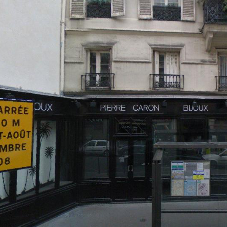}\\
(a) Input & (b) EC~\cite{nazeri-2019ICCVW-edgeconnect} &(c) GC~\cite{yu-2019iccv-freeform} &(d) PC~\cite{liu-2018eccv-particalcov}   & (e) RFR~\cite{Li_2020_CVPR} & (f) GT \\
\includegraphics[width=\swsix]{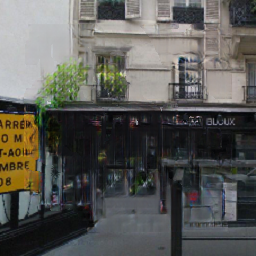}&
\includegraphics[width=\swsix]{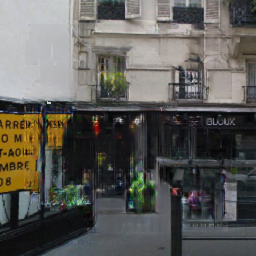}&
\includegraphics[width=\swsix]{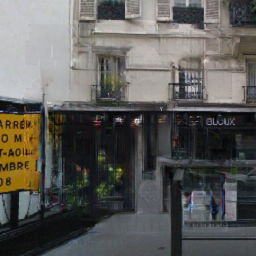}&
\includegraphics[width=\swsix]{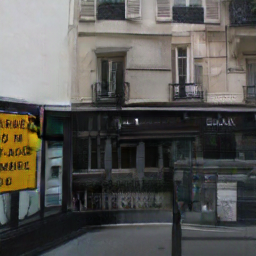}&
\includegraphics[width=\swsix]{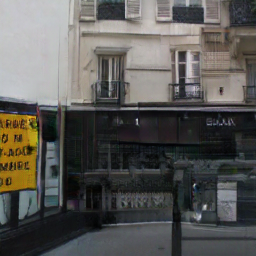}&
\includegraphics[width=\swsix]{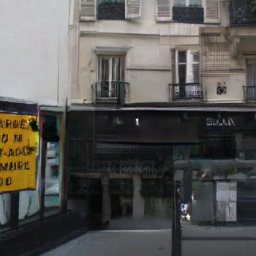}\\
(g) PIC$_1$~\cite{zheng2019pluralistic}  & (h) PIC$_2$  &(i) PIC$_3$  &(j) Ours$_1$   & (k) Ours$_2$  & (l) Ours$_3$  \\
\end{tabular}
\end{center}
\vspace{-2mm}
\caption{Qualitative comparisons with state-of-the-art methods on Paris Street View. Original images is in (f). Input images are in (a). The prior information is the output of PC in (d). The diverse outputs of PIC~\cite{zheng2019pluralistic} are in (g)-(i). The diverse outputs of our method are in (j)-(l).}
\label{fig:qualitativecity}
%\vspace{-2em}
\end{figure*}

\begin{figure*}[t]
    \begin{center}

\begin{tabular}{cccccc}
\vspace{-0.5mm}

\includegraphics[width=\swsix]{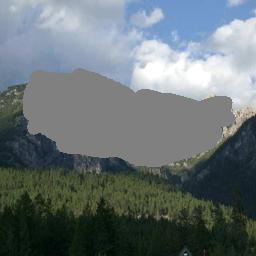}&
\includegraphics[width=\swsix]{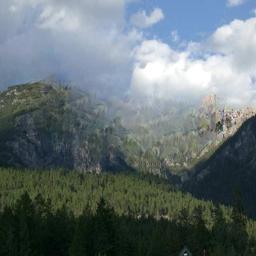}&
\includegraphics[width=\swsix]{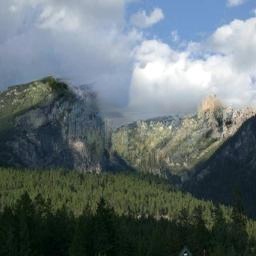}&
\includegraphics[width=\swsix]{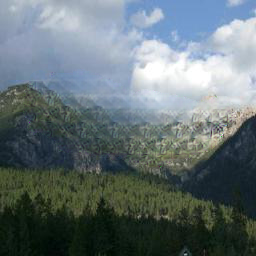}&
\includegraphics[width=\swsix]{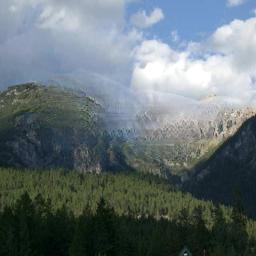}&
\includegraphics[width=\swsix]{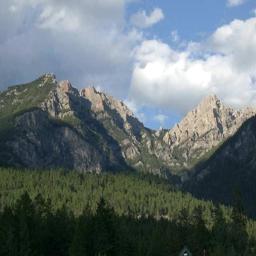}\\
(a) Input & (b) EC~\cite{nazeri-2019ICCVW-edgeconnect} &(c) GC~\cite{yu-2019iccv-freeform} &(d) PC~\cite{liu-2018eccv-particalcov}   & (e) RFR~\cite{Li_2020_CVPR} & (f) GT \\
\includegraphics[width=\swsix]{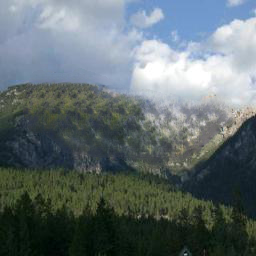}&
\includegraphics[width=\swsix]{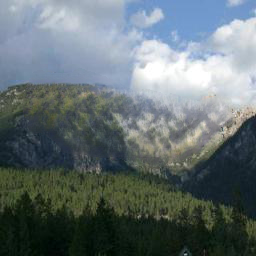}&
\includegraphics[width=\swsix]{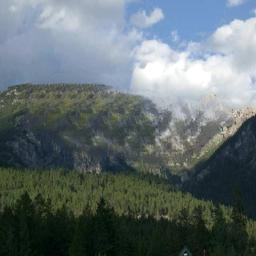}&
\includegraphics[width=\swsix]{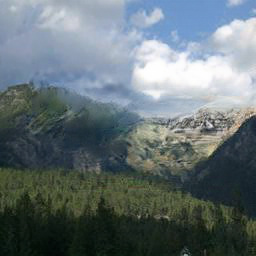}&
\includegraphics[width=\swsix]{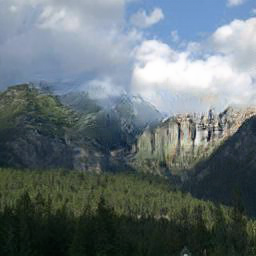}&
\includegraphics[width=\swsix]{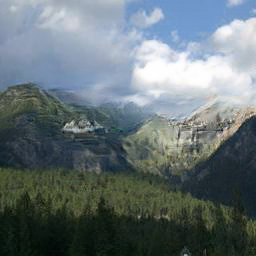}\\
(g) PIC$_1$~\cite{zheng2019pluralistic}  & (h) PIC$_2$  &(i) PIC$_3$  &(j) Ours$_1$   & (k) Ours$_2$  & (l) Ours$_3$  \\

\includegraphics[width=\swsix]{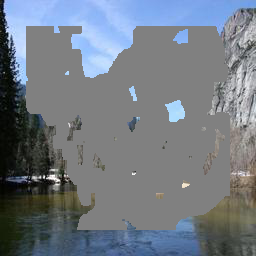}&
\includegraphics[width=\swsix]{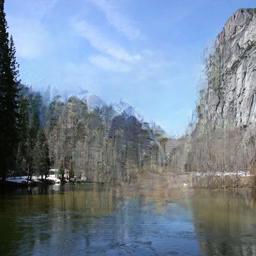}&
\includegraphics[width=\swsix]{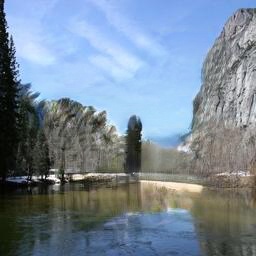}&
\includegraphics[width=\swsix]{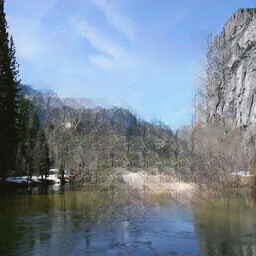}&
\includegraphics[width=\swsix]{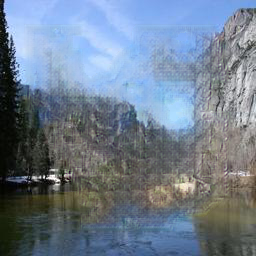}&
\includegraphics[width=\swsix]{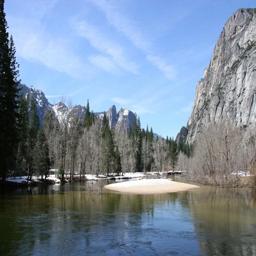}\\
(a) Input & (b) EC~\cite{nazeri-2019ICCVW-edgeconnect} &(c) GC~\cite{yu-2019iccv-freeform} &(d) PC~\cite{liu-2018eccv-particalcov}   & (e) RFR~\cite{Li_2020_CVPR} & (f) GT \\
\includegraphics[width=\swsix]{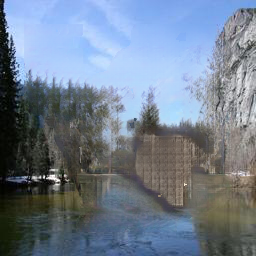}&
\includegraphics[width=\swsix]{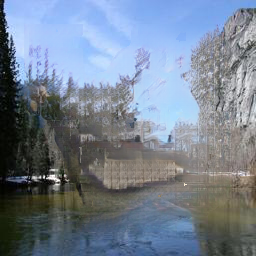}&
\includegraphics[width=\swsix]{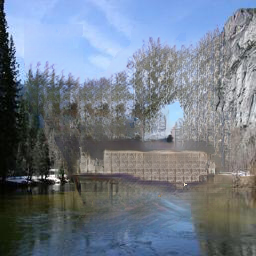}&
\includegraphics[width=\swsix]{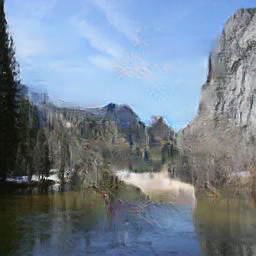}&
\includegraphics[width=\swsix]{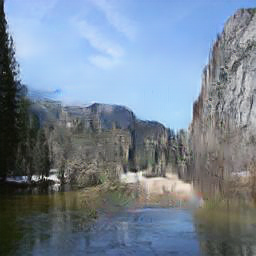}&
\includegraphics[width=\swsix]{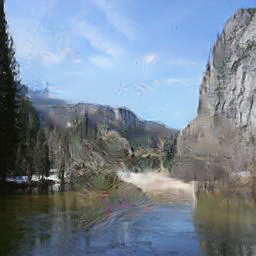}\\
(g) PIC$_1$~\cite{zheng2019pluralistic}  & (h) PIC$_2$  &(i) PIC$_3$  &(j) Ours$_1$   & (k) Ours$_2$  & (l) Ours$_3$  \\
\end{tabular}
\end{center}
\vspace{-2mm}
\caption{Qualitative comparisons with state-of-the-art methods on Place2. Original images is in (f). Input images are in (a). The prior information is the output of PC in (d).The diverse outputs of PIC~\cite{zheng2019pluralistic} are in (g)-(i). The diverse outputs of our method are in (j)-(l).}
\label{fig:qualitativenature}
%\vspace{-2em}
\end{figure*}

%  Proposed LOSS
\subsection{Perceptual Diversity Loss}
\label{sec:loss}

We make use of the same set of reconstruction losses between the generated image and the ground truth as in \cite{park2019semantic}. However, only minimizing reconstruction loss inhibits the diversity of results, since it essentially forces the model to learn a deterministic single mapping. \cite{mao2019mode} proposes a diversity loss to activate the diversity of generative models, which could be utilized in PG-GAN. We denote the generator of PD-GAN as $G$, and $I_{out1} = G(z_1,P,M)$ and  $I_{out2} = G(z_2,P,M)$ are two generated images conditioned on same coarse prediction $P$ and image mask $M$ but different latent vector $z_1$ and $z_2$. The diversity loss $L_{div}$ proposed in \cite{mao2019mode} is as follows:
\begin{equation}
  L_{div}= \frac { \lVert z_1-z_2 \rVert_1}{\lVert I_{out1}-I_{out2}\rVert_1 + \varepsilon},
\label{eq:codivers}
\end{equation}
which forces two output images to be farther in pixel space if their corresponding latent codes are far from each other. However, we find $L_{div}$ is not suitable for the diverse inpainting task. First of all, minimizing $L_{div}$ changes the content of the contextual regions that should be constant for different latent vectors. % Moreover, the final layer of PD-GAN is $Tanh$ which make the value of each point of result in the range of $-1~1$, so these value will close to 1 or -1 to maximize the $L_{cdiv}$ and promote the result to be all black or all white. To solve these issues, we propose a perceptual diversity loss $L_{div}$ which can be expressed as:
Moreover, we find the training very unstable. Minimizing $L_{div}$ promotes the results to be all black or all white in order to maximize the pixel distance between two output images. In this paper, we propose a simple but effective perceptual diversity loss $L_{pdiv}$ that tackles the above issues:
\begin{equation}
  L_{pdiv}= \frac { 1}{\sum_i  \lVert F_{i}( I_{out1})\cdot M-F_{i}( I_{out2})\cdot M \rVert_1  + \varepsilon} ,
\label{eq:pdivers}
\end{equation}
where $F_{i}$ is the $i$-th layer of a VGG-19 network \cite{simonyan2014very} pre-trained on ImageNet. In our work, $F_{i}$ corresponds to the activation maps from layers ReLU1\_1, ReLU2\_1, ReLU3\_1, ReLU4\_1, and ReLU5\_1. $L_{pdiv}$ keeps the context unchanged by introducing the mask in the loss. Meanwhile, the $L_{pdiv}$ is calculated on the perceptual space instead of raw pixel space. Maximizing the distance in highly non-linear network feature space integrates semantic measurement and avoids trivial solution that completely generates black or white pixels. Note that we do not involve the latent vectors in $L_{pdiv}$ as we want to maximize the perceptual distance of generated images no matter how close their latent vectors are. We find this further stabilizes the training.

In addition to the perceptual diversity loss, we follow the SPADE~\cite{park2019semantic} and utilize the reconstruction loss~\cite{Justin-eccv16-perceptual}, feature matching loss~\cite{wang2018high} and hinge adversarial loss~\cite{lim2017geometric} to optimize our network.

\begin{figure*}[t]
    \begin{center}
    \small
\begin{tabular}{cccccc}
\vspace{-0.5mm}
\includegraphics[width=\swsix]{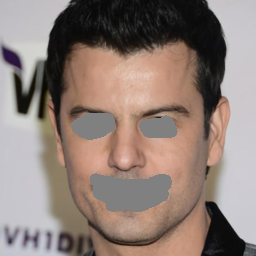}&
\includegraphics[width=\swsix]{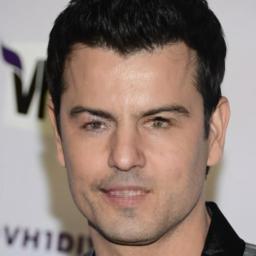}&
\includegraphics[width=\swsix]{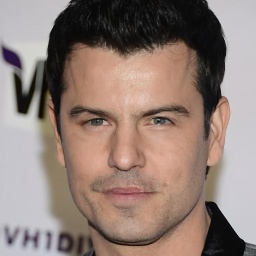}&
\includegraphics[width=\swsix]{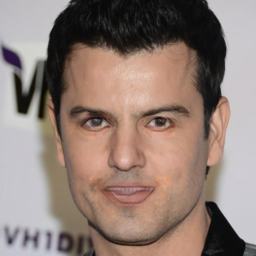}&
\includegraphics[width=\swsix]{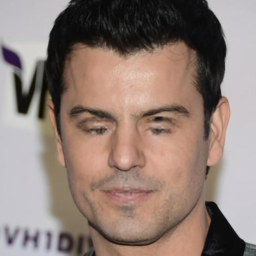}&
\includegraphics[width=\swsix]{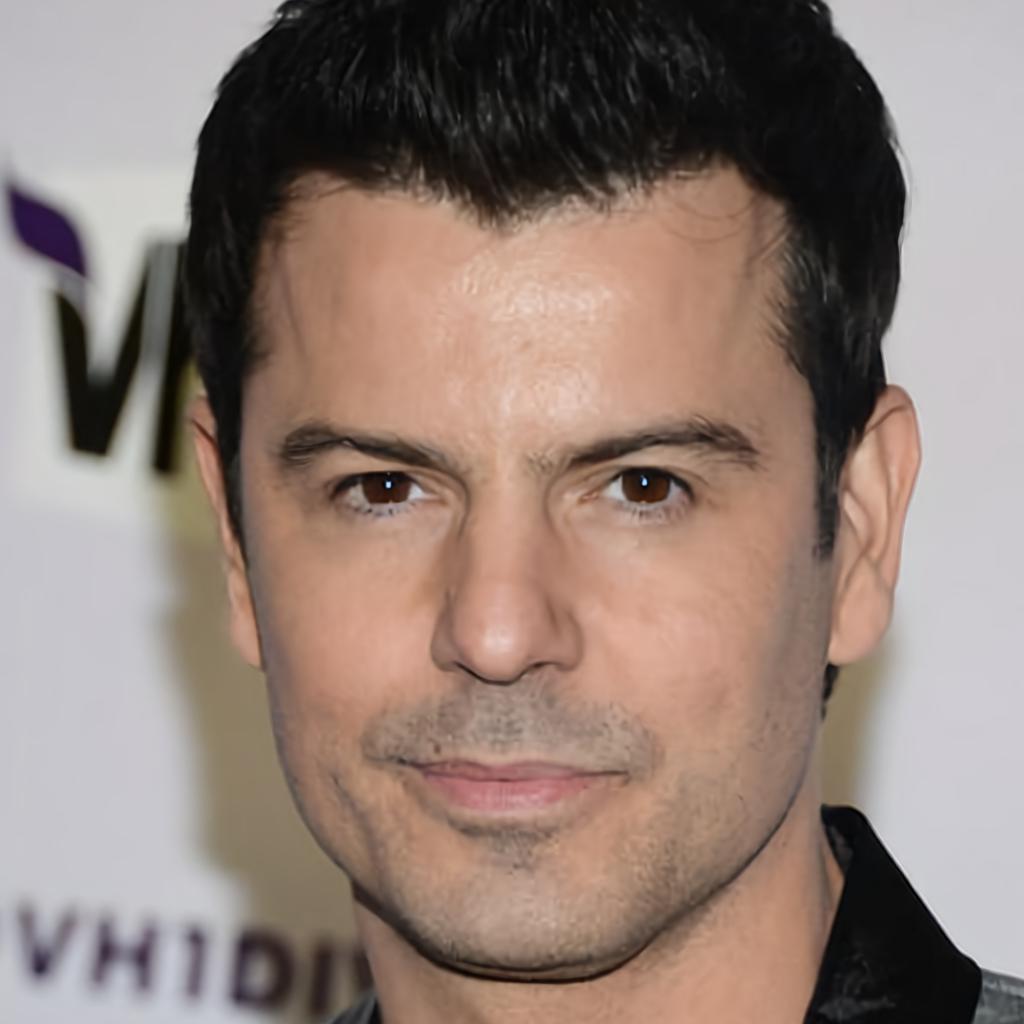}\\
(a) Input & (b) EC~\cite{nazeri-2019ICCVW-edgeconnect} &(c) GC~\cite{yu-2019iccv-freeform} &(d) PC~\cite{liu-2018eccv-particalcov}   & (e) RFR~\cite{Li_2020_CVPR} & (f) GT \\
\includegraphics[width=\swsix]{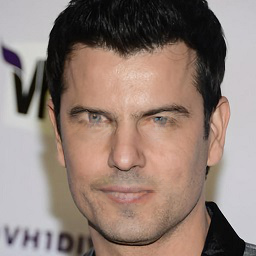}&
\includegraphics[width=\swsix]{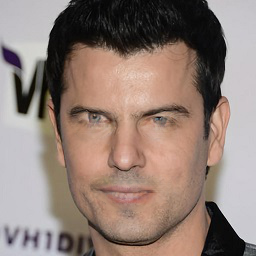}&
\includegraphics[width=\swsix]{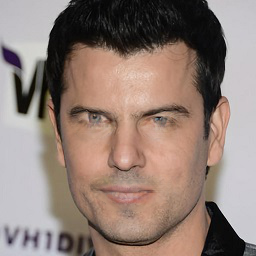}&
\includegraphics[width=\swsix]{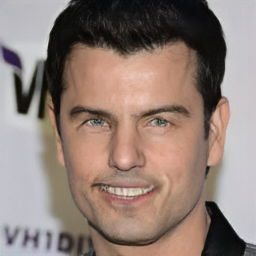}&
\includegraphics[width=\swsix]{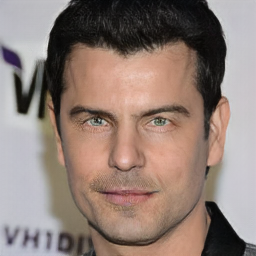}&
\includegraphics[width=\swsix]{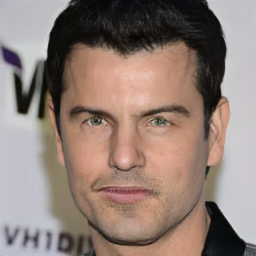}\\
(g) PIC$_1$~\cite{zheng2019pluralistic}  & (h) PIC$_2$  &(i) PIC$_3$  &(j) Ours$_1$   & (k) Ours$_2$  & (l) Ours$_3$  \\

\includegraphics[width=\swsix]{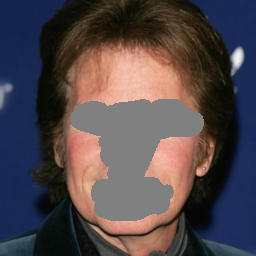}&
\includegraphics[width=\swsix]{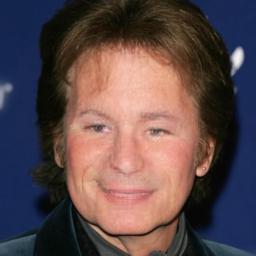}&
\includegraphics[width=\swsix]{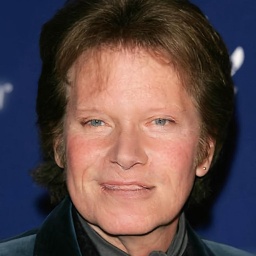}&
\includegraphics[width=\swsix]{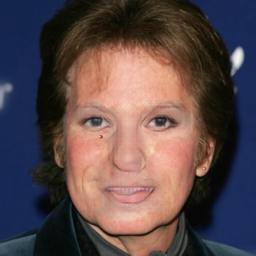}&
\includegraphics[width=\swsix]{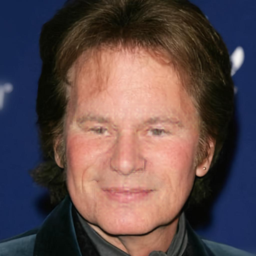}&
\includegraphics[width=\swsix]{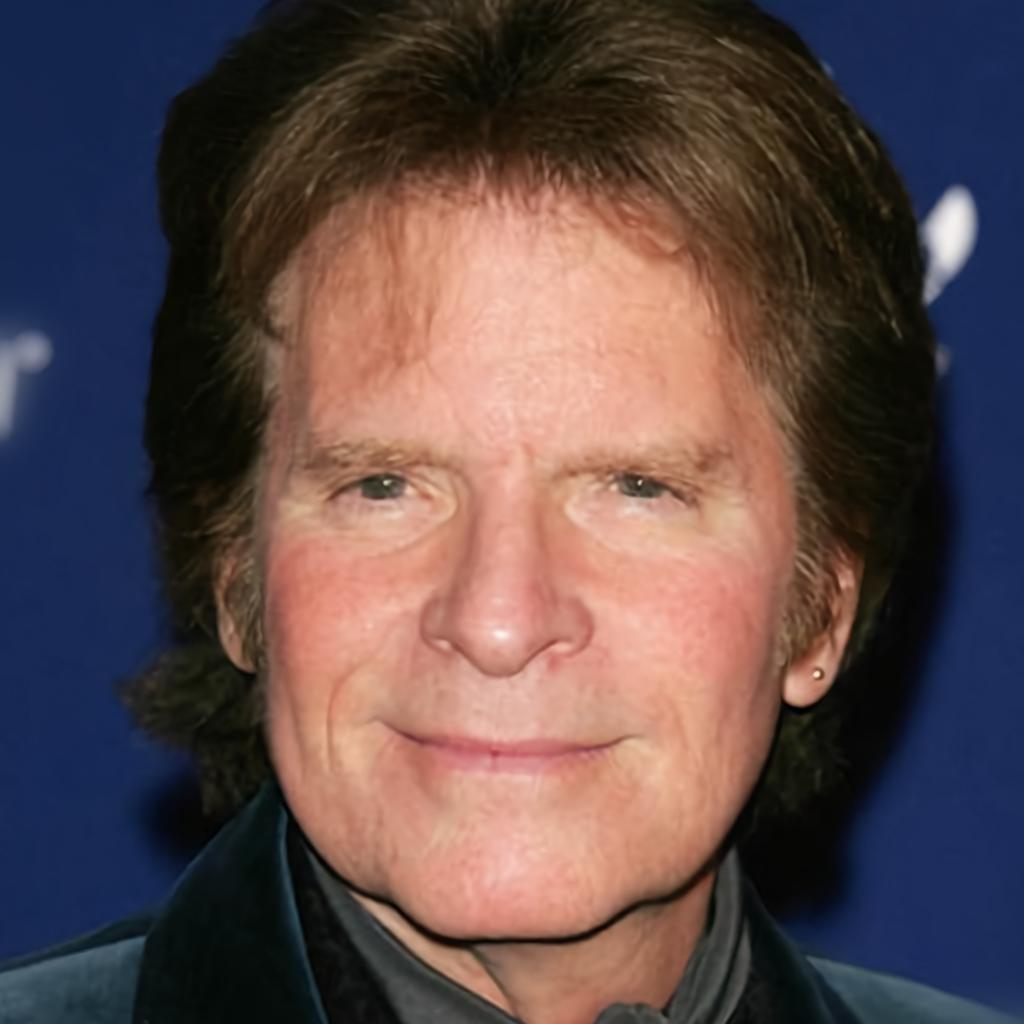}\\
(a) Input & (b) EC~\cite{nazeri-2019ICCVW-edgeconnect} &(c) GC~\cite{yu-2019iccv-freeform} &(d) PC~\cite{liu-2018eccv-particalcov}   & (e) RFR~\cite{Li_2020_CVPR} & (f) GT \\
\includegraphics[width=\swsix]{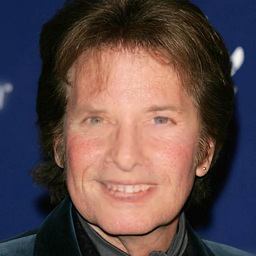}&
\includegraphics[width=\swsix]{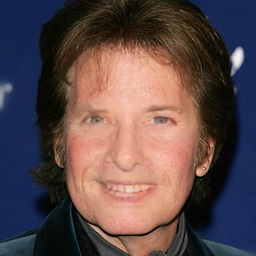}&
\includegraphics[width=\swsix]{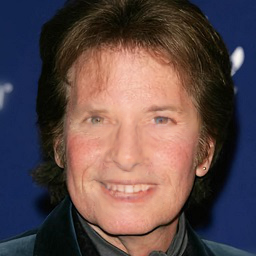}&
\includegraphics[width=\swsix]{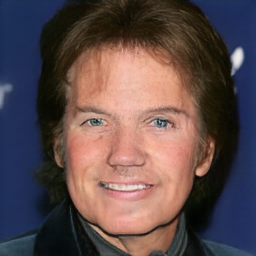}&
\includegraphics[width=\swsix]{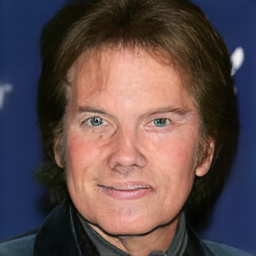}&
\includegraphics[width=\swsix]{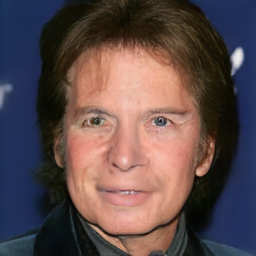}\\
(g) PIC$_1$~\cite{zheng2019pluralistic}  & (h) PIC$_2$  &(i) PIC$_3$  &(j) Ours$_1$   & (k) Ours$_2$  & (l) Ours$_3$  \\
\end{tabular}
\end{center}
\vspace{-2mm}
\caption{Qualitative comparisons with state-of-the-art methods on CelebA-HQ. Original images is in (f). Input images are in (a). The prior information is the output of PC in (d).The diverse outputs of PIC~\cite{zheng2019pluralistic} are in (g)-(i). The diverse outputs of our method are in (j)-(l).}
\label{fig:qualitativeface}
%\vspace{-2em}
\end{figure*}

\section{Experiments}
{\flushleft \bf Datasets.} We evaluated our proposed model on three datasets including Paris StreetView \cite{doersch-2015acm-makesparis}, CelebA-HQ~\cite{lee2020maskgan}, and Places2~\cite{Zhou-TPAMI17-place2su}. For the Paris StreetView~\cite{doersch-2015acm-makesparis} and CelebA-HQ~\cite{lee2020maskgan}, we use their original training and test splits. For the Places2~\cite{Zhou-TPAMI17-place2su}, we select the butte, canyon, field, moutain, mountain-path, mountain-snowy, sky, tundra and valley scene categories to train and validate our model following the shift-net~\cite{Yan-eccv18-shiftnet}. Since our model can generate multiple outputs, we randomly sampled 100 images for each masked image, and chose the top 5 results based on the discriminator scores for evaluation.

{\flushleft \bf Compared Methods.}  We compare with the following inpainting approaches: recurrent feature reasoning (RFR)~\cite{Li_2020_CVPR}, partial conv (PC)~\cite{liu-2018eccv-particalcov}, gated conv (GC)~\cite{yu-2019iccv-freeform}, edge connect (EC) ~\cite{nazeri-2019ICCVW-edgeconnect}, and PICNet (PIC)~\cite{zheng2019pluralistic}. Plus, we compare with CVAE~\cite{walker2016uncertain} and BicycleGAN~\cite{zhu2017toward} on the ability to generate diverse  results.

{\flushleft \bf Implementation Details.}
 All of our models are trained on irregular masks~\cite{liu-2018eccv-particalcov}. The mask and image are resized to  $256\times256$ as network input. We train the PC~\cite{liu-2018eccv-particalcov} following the official implementation. We set the inpainting results of the pre-trained PC~\cite{liu-2018eccv-particalcov} as the prior information. Our model is optimized using  Adam optimizer with $\beta_1$ = 0.0 and $\beta_2$  = 0.99. The initial learning rate is 1$\times$10$^{-4}$, and we utilize the TTUR~\cite{heusel-201nips-fid} strategy to train our model. We train the network for 500K iterations with batch size of 6. We choose low-dimensional manifold vector $|z|$ = 128 across all the datasets following SPADE~\cite{park2019semantic}.

 \renewcommand{\tabcolsep}{3pt}

\begin{table*}[t]
\centering
 \caption{Numerical comparisons on the Place2 dataset.  ${\downarrow}$ indicates lower is better while ${\uparrow}$ indicates higher is better.}
\begin{tabular}{c|cccc|cccc|cccc}
\hline
     & \multicolumn{4}{c|}{PSNR ${\uparrow}$}                                                                                                        & \multicolumn{4}{c|}{SSIM ${\uparrow}$}                                                                                                        & \multicolumn{4}{c}{FID ${\downarrow}$}                                                                                                         \\ \hline
Mask & 10-20\%                             & 20-30\%                             & 30-40\%                             & 40-50\%        & 10-20\%                             & 20-30\%                             & 30-40\%                             & 40-50\%        & 10-20\%                             & 20-30\%                             & 30-40\%                             & 40-50\%        \\ \hline
EC   & \multicolumn{1}{c|}{28.82}          & \multicolumn{1}{c|}{26.19}          & \multicolumn{1}{c|}{24.64}          & \textbf{23.29} & \multicolumn{1}{c|}{0.929}          & \multicolumn{1}{c|}{0.877}          & \multicolumn{1}{c|}{0.833}          & \textbf{0.783} & \multicolumn{1}{c|}{21.86}          & \multicolumn{1}{c|}{37.87}          & \multicolumn{1}{c|}{48.06}          & 59.27          \\ \hline
GC   & \multicolumn{1}{c|}{28.64}          & \multicolumn{1}{c|}{25.98}          & \multicolumn{1}{c|}{24.48}          & 22.96          & \multicolumn{1}{c|}{0.928}          & \multicolumn{1}{c|}{0.879}          & \multicolumn{1}{c|}{0.839}          & 0.784          & \multicolumn{1}{c|}{20.29}          & \multicolumn{1}{c|}{35.71}          & \multicolumn{1}{c|}{45.16}          & 56.07          \\ \hline
PC   & \multicolumn{1}{c|}{28.34}          & \multicolumn{1}{c|}{25.83}          & \multicolumn{1}{c|}{24.29}          & 22.58          & \multicolumn{1}{c|}{0.920}          & \multicolumn{1}{c|}{0.862}          & \multicolumn{1}{c|}{0.815}          & 0.772          & \multicolumn{1}{c|}{21.94}          & \multicolumn{1}{c|}{37.32}          & \multicolumn{1}{c|}{47.59}          & 59.29          \\ \hline
RFR   & \multicolumn{1}{c|}{29.16}          & \multicolumn{1}{c|}{26.38}          & \multicolumn{1}{c|}{24.19}          & 22.43          & \multicolumn{1}{c|}{0.923}          & \multicolumn{1}{c|}{0.859}          & \multicolumn{1}{c|}{0.774}          & 0.681          & \multicolumn{1}{c|}{21.62}          & \multicolumn{1}{c|}{36.72}          & \multicolumn{1}{c|}{59.00}          & 85.64          \\ \hline
PIC  & \multicolumn{1}{c|}{28.38}          & \multicolumn{1}{c|}{25.66}          & \multicolumn{1}{c|}{23.92}          & 22.46          & \multicolumn{1}{c|}{0.921}          & \multicolumn{1}{c|}{0.860}          & \multicolumn{1}{c|}{0.807}          & 0.744          & \multicolumn{1}{c|}{38.64}          & \multicolumn{1}{c|}{58.83}          & \multicolumn{1}{c|}{69.46}          & 99.17          \\ \hline
Ours & \multicolumn{1}{c|}{\textbf{29.20}} & \multicolumn{1}{c|}{\textbf{26.75}} & \multicolumn{1}{c|}{\textbf{25.48}} & 23.15          & \multicolumn{1}{c|}{\textbf{0.935}} & \multicolumn{1}{c|}{\textbf{0.880}} & \multicolumn{1}{c|}{\textbf{0.839}} & 0.782          & \multicolumn{1}{c|}{\textbf{19.98}} & \multicolumn{1}{c|}{\textbf{34.84}} & \multicolumn{1}{c|}{\textbf{44.24}} & \textbf{52.68} \\ \hline
\end{tabular}
\label{tab:Quantitative}
\end{table*}

\subsection{Comparison with Existing Work}

{\flushleft \bf Qualitative Comparisons.}
The qualitative comparisons on the results for filling irregular holes on Paris StreetView~\cite{doersch-2015acm-makesparis}, CeleBA-HQ~\cite{lee2020maskgan} and Place2~\cite{Zhou-TPAMI17-place2su}  are shown in Fig~\ref{fig:qualitativecity}, Fig~\ref{fig:qualitativeface} and Fig~\ref{fig:qualitativenature} respectively.  The result of PC in (d) is used as the prior information of our model.%The input images are shown in (a). The prediction of GC has distorted structure as shown in (c). he prior information is the output of PC in (d).

For the Paris StreetView, although EC, PC and RFR can generate roughly correct structure,  the results still contain blurry and unrealistic textures as shown in (b), (d) and (e), respectively. PIC is not effective to recover image content, and the results lack diversity. For CeleBA-HQ, the single-solution inpainting methods can generate natural but blurry content. The generated content of diverse image inpaitning method PIC is basically the same and seems blurry. For Place2, EC, PC and RFR get blurry and unnatural predictions. Although more visually pleasing content can be generated by GC as shown in (c), GC cannot produce diverse results. PIC can generate different content for a single masked input, however it lacks of obvious differences while  unreasonable semantics are rendered in the filling regions. In contrast to the above methods, our model generates multiple results with higher naturalness.
 \def\swsixa{0.080\linewidth}
\renewcommand{\tabcolsep}{0.5pt}
\begin{figure*}[t]
    \begin{center}

\begin{tabular}{ccccccccc}
\vspace{-0.5mm}

\includegraphics[width=\swsixa]{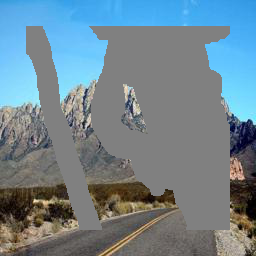}&
\includegraphics[width=\swsixa]{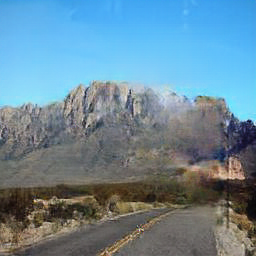}&
\includegraphics[width=\swsixa]{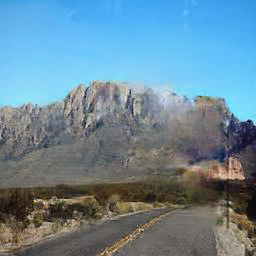}&
\includegraphics[width=\swsixa]{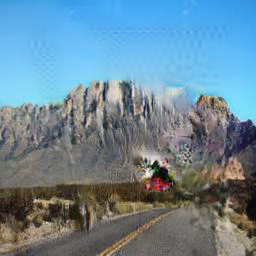}&
\includegraphics[width=\swsixa]{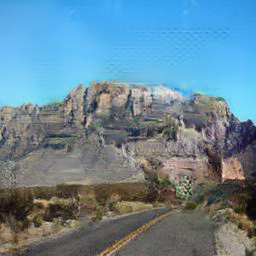}&
\includegraphics[width=\swsixa]{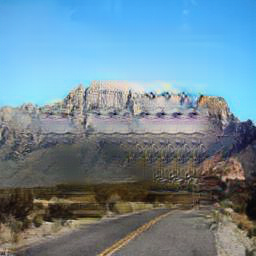}&
\includegraphics[width=\swsixa]{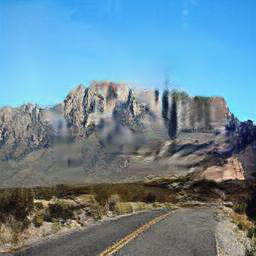}&
\includegraphics[width=\swsixa]{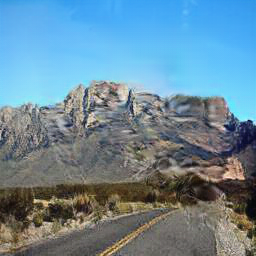}&
\includegraphics[width=\swsixa]{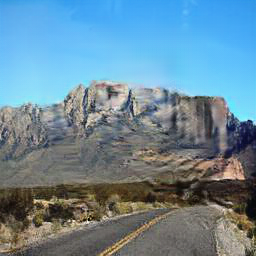}\\
Input & w/o diver$_1$    &w/o diver$_2$    &SPADE$_1$   & SPADE$_2$    &SPADE$_3$  &w/o h$_1$  & w/o h$_2$    &w/o h$_3$   \\

\includegraphics[width=\swsixa]{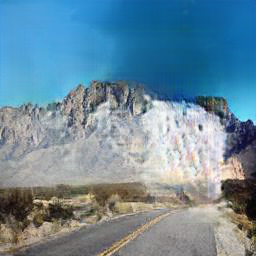}&
\includegraphics[width=\swsixa]{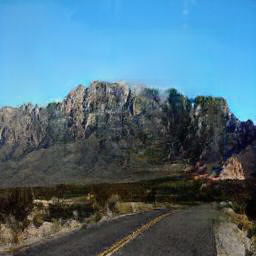}&
\includegraphics[width=\swsixa]{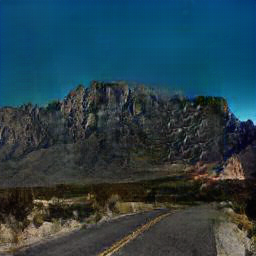}&
\includegraphics[width=\swsixa]{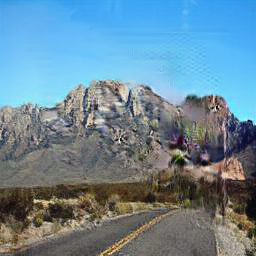}&
\includegraphics[width=\swsixa]{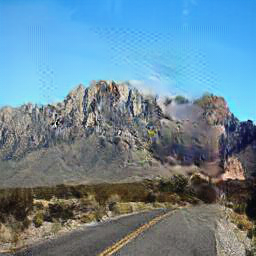}&
\includegraphics[width=\swsixa]{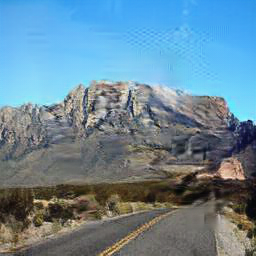}&
\includegraphics[width=\swsixa]{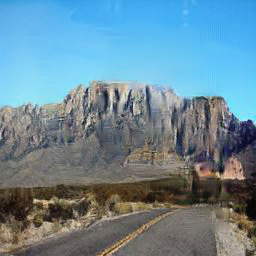}&
\includegraphics[width=\swsixa]{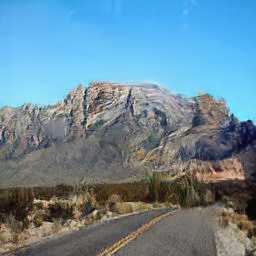}&
\includegraphics[width=\swsixa]{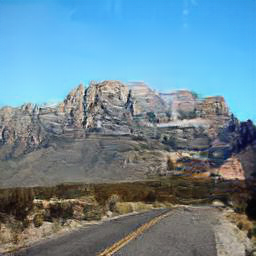}\\
w/ CDL$_1$    & w/ CDL$_2$     & w/ CDL$_3$  &w/o s$_1$  & w/o s$_2$    &w/o s$_3$    &Ours$_1$    & Ours$_2$     &Ours$_3$   \\

\end{tabular}
\end{center}
\vspace{-2mm}
\caption{ Visual results of ablation study. The nodiver$_{1\sim2}$ are the diverse results of our methods without diversity loss. The w/o h$_{1\sim3}$ are results after we replace the hard SPDNorm with soft.  The w/o s$_{1\sim3}$ are results after we replace the soft SPDNorm with hard. The predictions of our methods with conventional diversity loss~\cite{mao2019mode} are shown in CDL$_{1\sim3}$. The results of SPADE~\cite{park2019semantic} are in SPADE$_{1\sim3}$. Our results are in Ours$_{1\sim3}$.}
\label{fig:ablation}
%\vspace{-2em}
\end{figure*}

{\flushleft \bf Quantitative Comparisons.}
Since our model is used to solve the diverse image inpainting task, the generated results need to have both authenticity and diversity. We compare our method with baselines from two aspects: Realism and Diversity.
For realism comparison, our model is compared with both single- and diverse-solution inpainting methods on Place2. We follow PIC and assume that one of our top 5 samples
(ranked by the discriminator) will be close to the original ground truth, and select the single sample with the best balance of quantitative measures for comparison. For the evaluation metrics, we use the SSIM~\cite{wang-2004tip-ssim}, PSNR and FID~\cite{heusel-201nips-fid}. The evaluation results are shown in Table~\ref{tab:Quantitative}. For each hole versus image ratio, we randomly select 10 masks for testing.  Our method outperforms existing methods to fill irregular holes under various situations.

% \begin{table}[t]
% \centering
%  \caption{Quantitative comparison of diversity with the state of-the-art methods on the CelebA-HQ dataset.}
% \begin{tabular}{c|c|c|c|c}
% \hline
%      & \multicolumn{4}{c}{Diversity (LPIPS)${\uparrow}$}          \\ \hline
%   Method   & CVAE   & BicycleGAN & PIC    & Ours            \\ \hline
% $I_{out}$  & 0.0182 & 0.0436     & 0.0444 & \textbf{0.0472} \\ \hline
% $I_{out(m)}$ & 0.0251 & 0.0487     & 0.0493 & \textbf{0.0590} \\ \hline
% \end{tabular}
% \label{tab:diverse}
% \end{table}

\begin{table}[t]
\centering
 \caption{Quantitative comparison of diversity with the state of-the-art methods on the Places2 dataset.}
\begin{tabular}{c|c|c|c|c}
\hline
     & \multicolumn{4}{c}{Diversity (LPIPS)${\uparrow}$}          \\ \hline
  Method   & CVAE   & BicycleGAN & PIC    & Ours            \\ \hline
$I_{out}$  & 0.0432 & 0.0925     & 0.1096 & \textbf{0.1238} \\ \hline
$I_{out(m)}$ & 0.0506 & 0.1032     & 0.1547 & \textbf{0.1799} \\ \hline
\end{tabular}
\label{tab:diverse}
\end{table}

For the diversity comparison, we utilize the LPIPS metric~\cite{zhu2017toward, zheng2019pluralistic} to calculate the diversity score. The average score is calculated between 5K pairs generated from a sampling of 1K images of Places2 dataset. $I_{out}$ and $I_{out(m)}$ are the full image output and mask-region output, respectively. Our method obtains relatively higher diversity scores than other existing methods as shown in Table~\ref{tab:diverse}. To further demonstrate the superior of PD-GAN, we conduct extra user studies with 10 volunteers. Each subject is asked to compare 10 sets of inpainting results of PD-GAN and PIC and select the method with diverse. PD-GAN is favored in \textbf{83\%} of cases.

% \begin{table}[t]
% \centering
%  \caption{Quantitative comparison of ablation study on the CelebA-HQ dataset.}
% \begin{tabular}{c|c|c|c|c}
% \hline
%       & PSNR   ${\uparrow}$        & SSIM  ${\uparrow}$         & FID ${\downarrow}$           & LIPIS  ${\uparrow}$         \\ \hline
% w/ CDL  & 26.12          & 0.915          & 18.55          & 0.0490          \\ \hline
% w/o diver & \textbf{26.37} & \textbf{0.918} & 17.90          & 0.0172          \\ \hline
% w/o h  & 26.35          & 0.913          & 18.24          & 0.0516          \\ \hline
% w/o s  & 26.25          & 0.910          & 20.51          & \textbf{0.0623} \\ \hline
% SPADE  & 25.79          & 0.903          & 24.71          & 0.0479          \\ \hline
% Ours   & 26.32          & 0.915          & \textbf{16.83} & 0.0590          \\ \hline
% \end{tabular}
% \label{tab:ablation}
% \end{table}

\begin{table}[t]
\centering
 \caption{Quantitative ablation study on the Places2 dataset. ${\downarrow}$ indicates lower is better while ${\uparrow}$ indicates higher is better.}
\begin{tabular}{c|c|c|c|c}
\hline
       & PSNR   ${\uparrow}$        & SSIM  ${\uparrow}$         & FID ${\downarrow}$           & LIPIS  ${\uparrow}$         \\ \hline
w/ CDL  & 22.48         & 0.732         & 61.55          & 0.198          \\ \hline
w/o diver & \textbf{23.63} & \textbf{0.796} & 56.55         & 0.1934         \\ \hline
w/o h  & 23.03          & 0.771          & 61.22          & 0.2193         \\ \hline
w/o s  & 22.48         & 0.762          & 61.48          & \textbf{0.2215} \\ \hline
SPADE  & 22.26         & 0.779          & 60.55          & 0.198         \\ \hline
Ours   & 23.15          &0.782         & \textbf{52.68}   & 0.2206          \\ \hline
\end{tabular}
\label{tab:ablation}
\end{table}

\section{Ablation Study}
{\flushleft \bf SPDNorm.}
To evaluate the effects of SPDNorm, we compare the following ablations: 1) Replacing all the soft SPDNorm with hard SPDNorm (w/o s); 2) replacing all the hard SPDNorm with soft SPDNorm (w/o h); 3) Replacing all the SPDNorm with SPADE~\cite{park2019semantic}, which is equivalent to setting $D_h$ and $D_s$ to 1. Thus, SPADE can be regarded as a degenerated form of the proposed SPDNorm. As shown in Fig~\ref{fig:ablation},  SPADE makes training unstable and generates worse results, since SPADE unconditionally relies on the coarse prior information, which is contrary to the purpose of generating diverse results. The outputs of our method without soft SPDNorm (w/o s) has diverse details, but the artifacts are obvious. In comparison, the predictions of our method without hard SPDNorm (w/o h) contain meaningful content, but the diversity is declined.  By utilizing both soft and hard SPDNorm, our method achieves favorable results on both diversity  and quality. Table~\ref{tab:ablation} shows the similar numerical performance on the Places2 dataset where the combination of hard and soft SPDNorm is suitable for the diverse image inpainting task. We choose the mask ratio $40-50\%$ here.

{\flushleft \bf Perceptual Diversity Loss.}
We show the contributions of perceptual diversity loss (Eqn. \ref{eq:pdivers}) by removing it (w/o diver) or replacing it with conventional diversity loss~\cite{mao2019mode} in Eqn. \ref{eq:codivers} (w/ CDL). As shown in Fig~\ref{fig:ablation}, without using perceptual diversity loss the content generated lacks diversity. The content generated with the constraint of conventional diversity loss are more diverse. However, the recovered content tend to be all black or all white. The proposed perceptual diversity loss can solve the above issues. Similar performance has been shown numerically in Table~\ref{tab:ablation} where our full method achieves favorable results.

\section{Conclusion}
We propose a novel probabilistic diverse GAN (PD-GAN) for image inpainting. To get diverse inpainting results, PD-GAN utilizes prior information to modulate a random noise progressively. For the modulation process, PD-GAN adopts both soft and hard spatially probabilistic diversity normalization (SPDNorm) to control the probability of producing diverse results. Meanwhile, we propose the perceptual diversity loss to further boost diversity of PD-GAN. Experiments on a variety of datasets demonstrate that our PD-GAN cannot only produce diverse prediction, but also generates high-quality reconstruction content.

\clearpage
{\small
\bibliographystyle{ieee_fullname}
\bibliography{cvpr.bbl}
}

\end{document}